\documentclass[sigconf]{acmart}
\setcopyright{acmlicensed}
\copyrightyear{2026}
\acmYear{2026}
\acmDOI{XXXXXXX.XXXXXXX}
\acmISBN{}
\acmDOI{}

\acmConference[CARS@RecSys'26]{}{Sep 27 - Oct 2 2026}{Minnesota, USA}
\usepackage{booktabs}
\usepackage{multirow}
\usepackage{amsmath} % For \text command inside math mode
\usepackage{xcolor}  % Required for color styling
\usepackage{amsfonts} 
\definecolor{maroon}{rgb}{0.6, 0.0, 0.0}
\definecolor{forestgreen}{rgb}{0.0, 0.4, 0.2}

\newcommand\blfootnote[1]{%
  \begingroup
  \renewcommand\thefootnote{}\footnote{#1}%
  \addtocounter{footnote}{-1}%
  \endgroup
}

\begin{document}

\title{A Systematic Benchmark of Explainable Methods for Temporal Attribution in Sequential Recommendation Systems}

\author{Akash Pandey}
\orcid{0009-0003-5324-8667}
\email{akash.pandey@capitalone.com}
\affiliation{%
  \institution{Capital One, AI Foundations}
  \city{San Jose}
  \state{CA}
  \country{USA}
}

\author{Kanisha Shah}
\orcid{0009-0006-1031-3870}
\email{kanishaarpit.shah@capitalone.com}
\affiliation{%
  \institution{Capital One, AI Foundations}
  \city{New York}
  \state{NY}
  \country{USA}
}

\author{Addrish Roy}
\email{addrish.roy@capitalone.com}
\affiliation{%
  \institution{Capital One, AI Foundations}
  \city{New York}
  \state{NY}
  \country{USA}
}

\author{Dwipam Katariya}
\orcid{0009-0009-1058-1244}
\email{dwipam.katariya@capitalone.com}
\affiliation{%
  \institution{Capital One, AI Foundations}
  \city{McLean}
  \state{VA}
  \country{USA}
}

\author{Hongyangyang Shi}
\email{hongyangyang.shi@capitalone.com}
\affiliation{%
  \institution{Capital One, AI Foundations}
  \city{McLean}
  \state{VA}
  \country{USA}
}

\author{Amanda Ding}
\email{amanda.ding@capitalone.com}
\affiliation{%
  \institution{Capital One, AI Foundations}
  \city{San Jose}
  \state{CA}
  \country{USA}
}

\author{Kalanand Mishra}
\orcid{0000-0002-1832-1537}
\email{kalanand.mishra@capitalone.com}
\affiliation{%
  \institution{Capital One, AI Foundations}
  \city{San Jose}
  \state{CA}
  \country{USA}
}

\author{Pranab Mohanty}
\orcid{}
\email{pranab.mohanty@capitalone.com}
\affiliation{%
  \institution{Capital One, AI Foundations}
  \city{Seattle}
  \state{WA}
  \country{USA}
}

\renewcommand{\shortauthors}{Pandey et al.}

\begin{abstract}
  Sequential recommendation systems are central to modern personalization, exploiting user's historical interaction sequences to drive next-step decisions. Deep learning models,
  particularly convolutional and Transformer-based architectures, have proven highly effective at capturing temporal dependencies in these histories. For transparency and trust,
  understanding which past interactions drive a given recommendation is increasingly important — both for developers auditing model behavior and for users seeking a rationale.
  However, the non-linearities that give these models their predictive power also render them black boxes, making it difficult to attribute decisions to specific interactions.
  While gradient-based, perturbation-based, and attention-based explainability methods exist, a systematic benchmark of their faithfulness for sequential recommendation is
  missing. We address this gap by introducing a dual-model masking metric in which one model supplies per-timestep attribution scores and a separately trained, masking-robust
  probe measures the resulting change in predicted probability. Using this metric, we benchmark ten XAI methods across CNN, Transformer, SASRec, and BERT4Rec backbones on
  KuaiRand-1k and MovieLens-1M, complemented by analyses of temporal attribution patterns, item popularity confounding, and robustness to input corruption. Our key findings are:
  (i) gradient-based methods, particularly GradientSHAP and Integrated Gradients, yield the most faithful and robust attributions; (ii) raw attention weights are unreliable, but gradient-weighted attention restores faithfulness on shorter sequences, with degradation on longer horizons as softmax attention  probabilities converge toward uniform importance scores, diminishing the method's ability to identify informative interactions; and (iii) temporal attribution patterns in faithful methods reflect genuine task structure rather than recency or popularity bias.
\end{abstract}



\maketitle

\blfootnote{Copyright held by the author(s).}

\section{Introduction}
\label{sec:intro}

Sequential recommendation systems (SRS) underpin much of modern personalization, predicting a user's next action from the ordered sequence of past interactions rather than static preferences alone. Because the order, timing, and recency of interactions carry crucial contextual signals that non-sequential models discard, capturing these dynamic dependencies is central to recommendation quality. Deep architectures—particularly convolutional and self-attentive Transformer backbones—have become the workhorses for this task~\cite{hidasi2015session, tang2018personalized, kang2018sasrec, sun2019bert4rec, fintrec, timesync}, learning rich sequence representations that consistently outperform traditional models.

\begin{figure*}[t]
  \centering
  \includegraphics[width=0.8\textwidth]{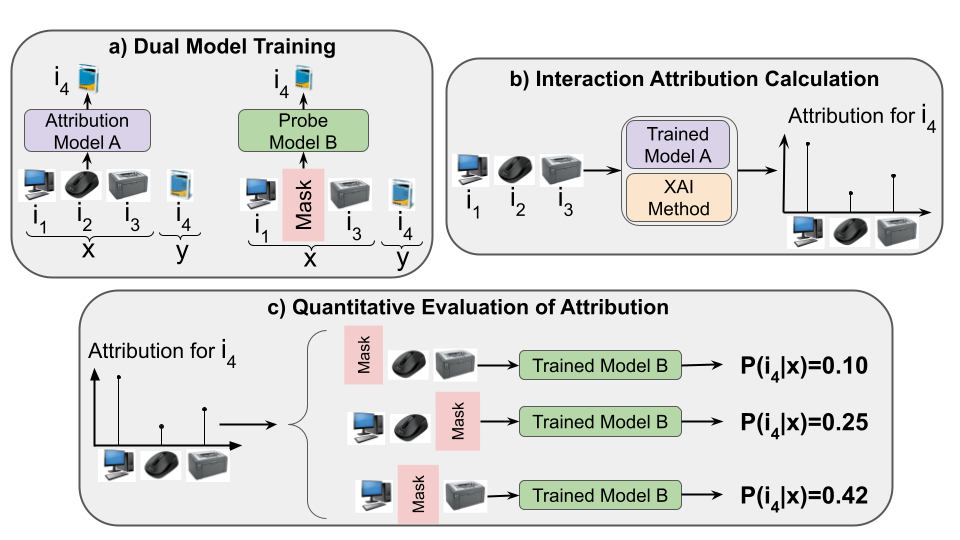}
  \vspace{-3mm}
  \caption{The faithfulness evaluation framework: (a) dual-model training of an attribution model (Model~$A$: \textit{Attribution model}) and a masked surrogate probe (Model~$B$: \textit{Probe model}); (b) per-interaction attribution computation via the target XAI method; and (c) quantitative faithfulness assessment verifying that the predicted probability from Model~$B$ drops proportionally when highly attributed interactions are masked.}
  \label{main_arch}
\end{figure*}

As SRS are deployed in higher-stakes settings such as financial services, digital content curation, and electronic commerce~\cite{spotify, alibaba, kuaiformer, pinnerformer, fintrec}, understanding whether a model genuinely exhibits contextual awareness has become as important as the recommendation itself. Temporal explainability provides a vital tool to evaluate this behavior. For developers, it helps audit and debug predictions~\cite{debug} by confirming that recommendations, such as suggesting action movies, stem from genuine historical preferences rather than recency or popularity bias. For users, explanations build trust and supply essential context~\cite{rudin2019, doshivelez2017, fintrec}, delivering transparent, personalized rationales like recommending a keyboard following a computer purchase. Yet, the stacked nonlinearities that grant these deep architectures their predictive power also render them black boxes, obscuring the mapping from individual past interactions to the final decision. A broad family of
post-hoc attribution methods has been proposed to address this, spanning gradient-based~\cite{ig, shap, deeplift}, perturbation-based~\cite{lime,
shap}, and attention-based approaches~\cite{attentiontracing, attention2}. While the correctness of these methods has been quantitatively evaluated in
computer vision (CV) and natural language processing (NLP) \cite{vitshap, gradsam}, a systematic assessment of their faithfulness for sequential recommendation remains absent. 
To address this gap, we define \emph{faithfulness} as the degree to which importance scores assigned to past interactions reflect the model's true
  decision process. As shown in Figure~\ref{main_arch}, we introduce a quantitative faithfulness metric based on a dual-model system: one model supplies per-timestep
  attribution scores, while a second architecturally identical model trained with stochastic interaction masking serves as a reusable in-distribution
  probe. This decoupling  avoids both the out-of-distribution problem of naive masking and the retraining cost of remove-and-retrain evaluations~\cite{hooker_2019, deyoung2020eraser}. In summary, this paper makes the following contributions:
  %\vspace{-1mm}
\begin{itemize}
  \item \textbf{Methodological Framework}:  To our knowledge, we are the first to introduce a dual-model faithfulness metric for temporal attribution in
  sequential recommendation, providing a practical, single-probe alternative to remove-and-retrain evaluation.
  \item \textbf{Systematic Benchmarking}: We evaluate ten gradient-, perturbation-, and attention-based attribution methods across CNN and Transformer
  backbones on KuaiRand-1k and MovieLens-1M datasets. 
  \item \textbf{Attribution Mechanics}:  We empirically examine whether recency or item popularity confounds attribution scores, finding that faithful
  methods reflect genuine task structure rather than sequence position or item frequency.
  \item \textbf{Robustness Diagnostics}:  We assess attribution stability under progressive input noise, identifying which methods remain consistent
  under realistic distribution shifts.
\end{itemize}
% \vspace{-1.5mm}
\textbf{Relevant Work:} Explainable recommendation has been studied across collaborative filtering, content-based, and knowledge graph models, with Zhang and Chen~\cite{zhang2020explainable} surveying explanation types from feature-based to social and visual approaches. In deep sequential models, attention weights have been widely used as a proxy for temporal importance~\cite{kang2018sasrec, sun2019bert4rec}, despite
  evidence that they are not always faithful~\cite{attention_not_working1, attention_not_working2, attention_not_working3, attention_not_working4}. Other work frames explainability as a joint learning objective, generating ranked justifications alongside recommendations~\cite{sce2021}, but such methods explain \emph{what} to recommend rather than \emph{which past interactions} drove a prediction. A complementary line of work applies counterfactual and causal reasoning to sequential models to disentangle genuine demand shifts from noise~\cite{coder}, offering interpretability through model design rather than post-hoc attribution. To the best of our knowledge, no prior work has systematically benchmarked post-hoc temporal attribution methods on sequential recommendation models. Our work
  fills this gap by evaluating ten XAI methods across multiple architectures and datasets along four axes: faithfulness, robustness, temporal attribution patterns, and popularity bias.

\section{Problem Formulation}
\label{sec:notation}

Let $\mathcal{U}$ and $\mathcal{I}$ denote the user and item sets, with $|\mathcal{U}|$ users and $|\mathcal{I}|$ items. For a user $u \in \mathcal{U}$, we observe a
chronologically ordered interaction sequence $S_u = (i_1, \dots, i_L)$ truncated to a fixed context window of length $L$. An embedding matrix $\mathbf{E} \in
\mathbb{R}^{|\mathcal{I}| \times d}$ maps each item to a $d$-dimensional vector $\mathbf{e}_t \in \mathbb{R}^d$, yielding the input representation
\begin{equation}
\mathbf{X} = \bigl[\mathbf{e}_1;, \mathbf{e}_2;, \dots;, \mathbf{e}_L\bigr] \in \mathbb{R}^{L \times d}.
\end{equation}
Depending on the dataset, $\mathbf{E}$ may encode raw item features (e.g., click counts, likes) or serve as a learnable lookup table trained end-to-end with the model.
A sequential recommendation model $f_{\theta}: \mathbb{R}^{L \times d} \to \mathbb{R}^{\gamma}$ maps the input sequence to a probability vector $\mathbf{y} \in  \mathbb{R}^{\gamma}$. For binary or multi-class prediction tasks (e.g., whether a user will engage with a given video), $\gamma = C$ where $C$ is the number of classes;
for next-item prediction, $\gamma = |\mathcal{I}|$.

%% Response to Dwipam: Do you need item embedings as well to compute the attribution score? Yes it basically defines your input for the model f_theta
A post-hoc explanation method $g(f_\theta, \mathbf{X}, c) \to \boldsymbol{\phi} \in \mathbb{R}^L$ assigns a scalar attribution score $\phi_j$ to each position (user interaction) $j \in
[1, L]$ in the sequence with respect to a target class or item $c$. A positive $\phi_j$ indicates that the interaction at time step $j$ increases the predicted
probability $y_c$, while a negative $\phi_j$ indicates that it suppresses it.

Given the interaction attribution scores $\boldsymbol{\phi}$ defined in Section~\ref{sec:notation}, this work systematically evaluates their quality
along three axes: \textbf{faithfulness} — whether $\boldsymbol{\phi}$ accurately reflects the model's true decision process; \textbf{robustness} —
whether $\boldsymbol{\phi}$ remains stable under progressive input noise and realistic distribution shifts; and \textbf{attribution mechanics} —
whether $\boldsymbol{\phi}$ is confounded by recency or item popularity rather than reflecting genuine task structure and user preference signals.

\section{Methodology}
\label{sec:method}
 
We introduce quantitative metrics to benchmark post-hoc XAI methods for sequential recommendation, covering faithfulness via dual-model masking, robustness under input noise, and popularity bias assessment. 

% %%%% Response to Dwipam: \dwipam{is this with attention-based mask or item embedding whitening? - } yes we are just padding the inputs randomly.....
% \noindent\textbf{Interaction Attribution Score Calculation.} We benchmark ten XAI methods on their effectiveness in estimating interaction attribution scores $\boldsymbol{\phi}$ (Section~\ref{sec:notation}): gradient-based methods including Integrated Gradients~\cite{ig}, GradientSHAP~\cite{shap}, DeepLIFT~\cite{deeplift}, DeepLIFT-SHAP~\cite{shap}, and Grad-CAM~\cite{gradcam}; perturbation-based
%   methods including LIME~\cite{lime} and KernelSHAP~\cite{shap}; attention-based methods including Attention Tracing~\cite{attention2, attentiontracing} and Grad-SAM~\cite{gradsam}; and the intrinsically interpretable TimeSliver~\cite{pandey2026timesliver}. For implementation details and method descriptions, we refer the reader to the cited publications and Appendix~\ref{app:xai_impl}.

\noindent \textbf{Faithfulness via Dual-Model Masking.} To avoid the retraining cost and instability of ROAR~\cite{hooker_2019}, we adopt a dual-model strategy (Figure \ref{main_arch}a) that decouples the model being \emph{explained} from the model being \emph{probed}.
% \vspace{-3mm}

\noindent\textbf{\textit{Attribution and probe models.}}
We train two variants of each backbone architecture. Model~$A$ ($\theta_A$) is trained conventionally on complete sequences and supplies all  attribution vectors $\boldsymbol{\phi}(f_{\theta_A}, \mathbf{X})$ as shown in Figure \ref{main_arch}b. Model~$B$ ($\theta_B$, identical architecture) is trained with stochastic
  interaction masking, where each timestep is independently replaced by $\mathbf{m}=\mathbf{0}$ with probability $p$, so that $B$ learns to predict
  under incomplete histories. We set $p=0.05$ by default; this is a dataset-dependent hyperparameter and if changed, Model~$B$ must be retrained
  accordingly. At evaluation time, we probe using Model~$B$ since interactions are masked with the same probability $p$ seen during its training,
  ensuring masked inputs remain in-distribution and directly addressing the out-of-distribution problem that makes standard perturbation-based
  faithfulness metrics unreliable~\cite{fong2017interpretable, samek2016evaluating}.
% \vspace{-2mm}

\noindent\textbf{\textit{Faithfulness Accounting.}}
Faithfulness measures the degree to which an attribution method's assigned importance scores accurately reflect the sequential model's true reliance on specific historical interactions when making a prediction. To quantify this alignment, we evaluate how the model's output probability degrades as highly attributed interaction timesteps are systematically removed. Formally, given a sequence $\mathbf{X}$, we sort all $L$ timesteps by their attribution score $\phi_t$ calculated using Model $A$ in descending order and partition them into $G = \lceil L/k \rceil$ non-overlapping groups of size $k = \lfloor 0.05 \times L \rfloor$,

% \akash{Faithfulness measures the degree to which an attribution method's assigned importance scores accurately reflect the sequential model's true reliance on specific historical interactions when making a prediction.}
%   Given a sequence $\mathbf{X}$, we sort all $L$ timesteps by their attribution score $\phi_t$ calculated using Model $A$ in descending order and
%   partition them into $G = \lceil L/k \rceil$ non-overlapping groups of size $k = \lfloor 0.05 \times L \rfloor$,
  \begin{equation}
    \mathcal{G}_1, \dots, \mathcal{G}_G, \qquad
    t \in \mathcal{G}_g,\; t' \in \mathcal{G}_{g+1}
    \;\Rightarrow\; \phi_t \geq \phi_{t'},
  \end{equation}
  so that $\mathcal{G}_1$ contains the $k$ highest-attributed timesteps and $\mathcal{G}_G$ the lowest. Let $M_{\mathcal{G}_g}(\mathbf{X})$ denote the
  masked sequence obtained by replacing the embeddings at all positions in $\mathcal{G}_g$ with $\mathbf{m}$, while leaving all other positions
  unchanged. For each group $g$, we form a pair $(\bar{\phi}_g, y_g)$, where $\bar{\phi}_g = \frac{1}{|\mathcal{G}_g|} \sum_{t \in \mathcal{G}_g}
  \phi_t$ is the mean attribution within the group, and $y_g = \frac{1}{|\mathcal{D}|} \sum_{\mathbf{X} \in \mathcal{D}}
  f_{\theta_B}\!\bigl(M_{\mathcal{G}_g}(\mathbf{X})\bigr)_c$ is the mean predicted probability of the target class $c$ by Model~$B$ when that group is
  masked, averaged over all test sequences. A faithful method should rank highly the groups whose removal causes the largest drop in $y_g$ (Figure~\ref{main_arch}c). We quantify faithfulness as the Pearson correlation~\cite{pcc} $c_f$ between the $G$ pairs $\{(\bar{\phi}_g, y_g)\}_{g=1}^{G}$,
   where $\overline{\phi} = \frac{1}{G}\sum_g \bar{\phi}_g$ and $\bar{y} = \frac{1}{G}\sum_g y_g$. A more negative $c_f$ (closer to $-1$) indicates that
   the attribution method more faithfully reflects the model's reliance on each part of the interaction history, as masking highly attributed groups
  causes the largest drop in predicted probability.

\noindent \textbf{Robustness to Input Noise.} Reliable attribution methods should produce stable importance scores under noisy or incomplete interaction histories, a common occurrence in
  real-world recommender systems due to logging errors, missing data, or distribution shifts. For a sequence $\mathbf{X}$, we randomly zero out a
  fraction $\eta \in \{0.05, 0.10, 0.15, 0.20, 0.25\}$ of timesteps to obtain a corrupted sequence $\mathbf{X}^{(\eta)}$, recompute
  $\boldsymbol{\phi}^{(\eta)}$ using Model~$A$, and measure the Pearson correlation~\cite{pcc} $c_r(\eta)$ between clean and corrupted attributions over
   unmasked positions $t \notin \mathcal{M}^{(\eta)}$, averaged over all test sequences and multiple random seeds. A method with $c_r(\eta)$ close to
  $1$ across all noise levels is considered robust.

\noindent \textbf{Estimating Popularity Bias.} Let $\text{freq}(i)$ denote the number of times item $i$ appears in $\mathcal{D}_{\text{train}}$, and let $\bar{\phi}(i)$ denote its mean attribution  score across all positions in $\mathcal{D}_{\text{test}}$ where it appears. We compute the Pearson correlation $c_p$ between $\{\text{freq}(i)\}$ and  $\{\bar{\phi}(i)\}$ over all items. While popular items may carry genuine predictive signal, an excessively high $c_p$ suggests that a method
  attributes importance based on global item frequency rather than localised user interaction context, allowing us to evaluate the extent to which
  attribution methods are confounded by popularity profiles.

\subsection{Implementations Details}
\label{sec:impl}
\noindent\textbf{Architecture Selection.} We train five backbone architectures. \textit{CNN} captures local temporal interactions through convolutional filters and has been applied to sequential recommendation~\cite{cnn1, cnn2}. \textit{Vanilla Transformer}~\cite{vaswani2017attention, transformer2} leverages global self-attention to
   model long-range dependencies across interaction histories, making it a strong general-purpose backbone for sequential modelling.
  \textit{SASRec}~\cite{kang2018sasrec} and \textit{BERT4Rec}~\cite{sun2019bert4rec} are widely adopted sequential recommenders included for their
  practical relevance. \textit{TimeSliver}~\cite{pandey2026timesliver} is an intrinsically interpretable model recently shown to produce faithful
  temporal attributions, serving as an interpretable reference point in our benchmark.

\noindent\textbf{Baseline XAI Methods.} We benchmark ten XAI methods on their effectiveness in estimating interaction attribution scores $\boldsymbol{\phi}$ (Section~\ref{sec:notation}): gradient-based methods including Integrated Gradients~\cite{ig}, GradientSHAP~\cite{shap}, DeepLIFT~\cite{deeplift}, DeepLIFT-SHAP~\cite{shap}, and Grad-CAM~\cite{gradcam}; perturbation-based
  methods including LIME~\cite{lime} and KernelSHAP~\cite{shap}; attention-based methods including Attention Tracing~\cite{attention2, attentiontracing} and Grad-SAM~\cite{gradsam}; and the intrinsically interpretable TimeSliver~\cite{pandey2026timesliver}. For implementation details and method descriptions, we refer the reader to Appendix~\ref{app:impl}.

 \begin{table*}[!htbp]
        \centering
        \caption{Faithfulness score $c_f$ on KuaiRand-1k (binary classification) and MovieLens-1M (next-item prediction).
    \textbf{Higher magnitude
      negative values indicate stronger faithfulness}. The best methods are highlighted in \textcolor{forestgreen}{\textbf{dark forest green}} (strongest
      negative correlation), while the worst-performing methods are highlighted in \textcolor{maroon}{\textbf{maroon}} (unfaithful positive or near-zero
      correlation).}
        \vspace{-3mm}
        \label{tab:faithfulness}
        \footnotesize
        \setlength{\tabcolsep}{4pt}
        \renewcommand{\arraystretch}{0.5}
        \resizebox{\textwidth}{!}{%
        \begin{tabular}{lllccc}
        \toprule
        \textbf{Architecture} & \textbf{Variant} & \textbf{Method}
            & \textbf{KuaiRand-1k}
            & \textbf{KuaiRand-1k-Long}
            & \textbf{MovieLens-1M} \\
        \midrule
        Baseline & & Random & $-0.071 \pm 0.214$ & $+0.0644 \pm 0.1427$ & $-0.012 \pm 0.145$ \\
        \midrule
        \multirow{7}{*}{CNN} & \multirow{7}{*}{---}
            & Integrated Gradients  & $-0.972 \pm 0.010$ & $-0.9084 \pm 0.0221$ & $-0.952 \pm 0.008$ \\
            && GradientSHAP         & \textcolor{forestgreen}{\textbf{$-0.986 \pm 0.003$}} & $-0.9401 \pm 0.0142$ & $-0.792 \pm
  0.020$ \\
            && DeepLift             & $-0.976 \pm 0.007$ & $-0.9337 \pm 0.0150$ & $-0.799 \pm 0.022$ \\
            && DeepLiftSHAP         & $-0.976 \pm 0.007$ & $-0.9337 \pm 0.0150$ & $-0.799 \pm 0.022$ \\
            && KernelSHAP           & $-0.978 \pm 0.005$ & $-0.8232 \pm 0.0714$ & $-0.867 \pm 0.007$ \\
            && LIME                 & $-0.582 \pm 0.207$ & $-0.2236 \pm 0.1674$ & $-0.897 \pm 0.007$ \\
            && Grad-CAM             & $+0.067 \pm 0.282$ & \textcolor{maroon}{$+0.4576 \pm 0.2905$} & \textcolor{maroon}{$+0.021 \pm 0.022$} \\
        \midrule
        \multirow{6}{*}{Transformer}
            & \multirow{2}{*}{Vanilla}
                & Attention Tracing  & \textcolor{maroon}{$+0.930 \pm 0.043$} & $+0.1781 \pm 0.0812$ & $-0.951 \pm 0.007$ \\
            && Grad-SAM             & $-0.825 \pm 0.055$ & $-0.3077 \pm 0.1546$ & \textcolor{forestgreen}{\textbf{$-0.986 \pm 0.004$}} \\
            \cmidrule{2-6}
            & \multirow{2}{*}{SASRec}
                & Attention Tracing  & $+0.375 \pm 0.0465$ & $+0.142 \pm 0.1777$ & $-0.032 \pm 0.053$ \\
            && Grad-SAM             & $-0.948 \pm 0.0124$ & \textcolor{forestgreen}{$-0.983 \pm 0.0047$} & $-0.904 \pm 0.016$ \\
            \cmidrule{2-6}
            & \multirow{2}{*}{BERT4Rec}
                & Attention Tracing  & $-0.746 \pm 0.066$ & $-0.529 \pm 0.4149$ & $-0.944 \pm 0.005$ \\
            && Grad-SAM             & $-0.881 \pm 0.0497$ & $-0.9232 \pm 0.0438$ & $-0.961 \pm 0.005$ \\
        \midrule
        TimeSliver & --- & TimeSliver & $-0.961 \pm 0.024$ & $-0.8877 \pm 0.0834$ & $-0.841 \pm 0.003$ \\
        \bottomrule
        \end{tabular}}
    \label{main_table}
    \end{table*}
\vspace{-3mm}
\section{Results and Discussions}
\label{sec:results}

% We organize our findings around three questions: faithfulness (Section~\ref{sec:res-quant}), robustness under input noise (Section~\ref{sec:res-robust}), and attribution mechanics covering recency and popularity bias (Section~\ref{sec:res-mechanics}).

We organize our findings around three questions: faithfulness (Section~\ref{sec:res-quant}), robustness under input noise (Section~\ref{sec:res-robust}), and attribution mechanics covering recency and popularity bias (Section~\ref{sec:res-mechanics}).
% \vspace{-2mm}
% \vspace{-5mm}
\subsection{Datasets}
\label{datasets}

We use two public datasets: KuaiRand-1k~\cite{gao2022kuairand} and MovieLens-1M~\cite{harper2015movielens}, each split into
  $\mathcal{D}_{\text{train}}$, $\mathcal{D}_{\text{valid}}$, and $\mathcal{D}_{\text{test}}$; all evaluations are performed on
  $\mathcal{D}_{\text{test}}$.  KuaiRand-1k is a short-video interaction dataset where the task is binary classification, specifically predicting whether a user will long-view the  most recent video, defined as a watch duration of at least 18,000 ms~\cite{gao2022kuairand}. Since the primary goal is to benchmark explanation  methods rather than predictive performance, we downsample from 1,000 to 100 users to keep experimental cost tractable, validated by AUC-ROC scores on  the 100-user subset that are comparable to reported values in the literature~\cite{nature_kuairand} (Table~\ref{tab:predictive}), confirming that the
  models capture representative interaction patterns sufficient for attribution benchmarking.
 We evaluate under two  sequence lengths, $L=100$ and $L=1000$, to study whether explainability insights change over longer interaction horizons. MovieLens-1M is a movie  rating dataset with a next-item prediction task, evaluated at $L=200$ following prior work~\cite{kang2018sasrec, sun2019bert4rec}. Key statistics are
  reported in Table~\ref{tab:datasets} in the Appendix.

\noindent\textbf{Training details.} KuaiRand-1k models use cross-entropy loss on a class-balanced subset; MovieLens-1M models on positive and negative
   samples. Early stopping on validation performance is used in both cases. All models are trained with Adam (lr=$0.001$). Gradient-based and
  perturbation-based methods are implemented via Captum~\cite{kokhlikyan2020captum}. Full architecture, XAI configuration, and reproducibility details
  are provided in Appendix~\ref{app:impl}.
  
  \begin{figure*}[t]
    \centering
    \includegraphics[width=0.9\linewidth]{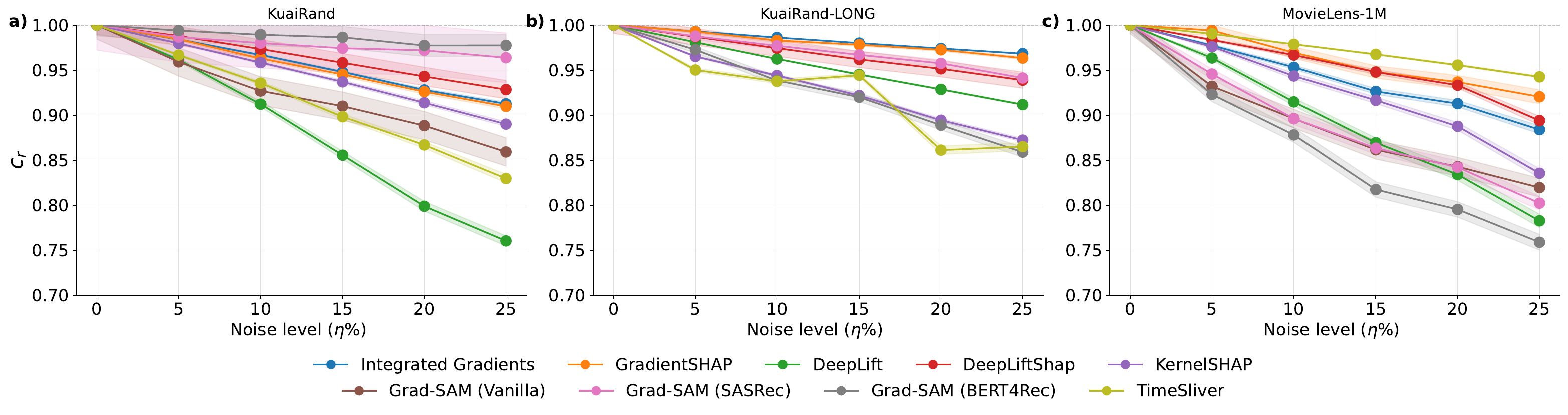}
    % \vspace{-4mm}
    \caption{Robustness score $c_r(\eta)$ between clean and noise-corrupted importance scores, averaged over 200 faithful samples across a) KuaiRand-1k,
  b) KuaiRand-1k-LONG, and c) MovieLens-1M.}
    \label{fig:robustness}                                                                             
  \end{figure*}  

\subsection{Faithfulness Evaluation of XAI methods}
\label{sec:res-quant}

Table~\ref{main_table} reports $c_f$ across all methods and datasets. Integrated Gradients and Grad-SAM applied on SASRec and BERT4Rec are the most reliably faithful methods,
  achieving $c_f < -0.9$ consistently across all settings. GradientSHAP, DeepLift, and DeepLiftSHAP are also highly faithful with $c_f < -0.8$ across both KuaiRand-1k settings,
  confirming that gradient-based attribution is well-suited for sequential recommendation. TimeSliver achieves $c_f < -0.8$ consistently across all datasets, which is notable
  given that it is a unified predictive and explainable model with up to $5\times$ fewer parameters than CNN and Transformer backbones (Table~\ref{app_tab:num_parameters}),
  demonstrating that intrinsic interpretability can match the best post-hoc methods in faithfulness.

Attention Tracing is highly inconsistent and architecture-dependent: it is actively anti-faithful on KuaiRand-1k for the Vanilla Transformer ($c_f = +0.930$), while improving on
   BERT4Rec ($c_f = -0.746$ on KuaiRand-1k, $-0.944$ on MovieLens-1M), yet consistently remaining below Grad-SAM across all settings. Grad-SAM on SASRec and BERT4Rec achieves $c_f < -0.9$ on KuaiRand-1k and MovieLens-1M; however, Grad-SAM (Vanilla) degrades substantially on KuaiRand-1k-Long ($c_f = -0.308$), exposing a vulnerability to longer interaction horizons where softmax attention
  increasingly converges toward uniform token selection~\cite{mudarisov2025limitations}, with performance falling for $L > 300$ as shown in
  Table~\ref{app:tab:gradsam_length}.

% \vspace{-5mm}
\subsection{Robustness of XAI methods}
\label{sec:res-robust}

We assess robustness using $c_r(\eta)$ (Section~\ref{sec:method}), restricting our analysis to methods proven faithful under $c_f$ (Table~\ref{main_table}). As shown in Figure~\ref{fig:robustness}, no method degrades by more than 25\% under progressive noise, indicating that faithfulness and robustness are largely complementary properties in sequential recommendation. Among the evaluated methods, DeepLift, KernelSHAP, and TimeSliver exhibit comparatively higher attribution degradation under noise on the KuaiRand datasets, while DeepLift and Grad-SAM (on all Transformer variants) are most sensitive to input corruption across all Transformer variants on MovieLens-1M. Conversely, Integrated Gradients, GradientSHAP, and DeepLiftSHAP consistently show the least degradation across all settings, reinforcing their suitability as stable XAI methods for sequential recommendation.

\begin{figure*}[t]
  \centering
 \includegraphics[width=0.8\linewidth]{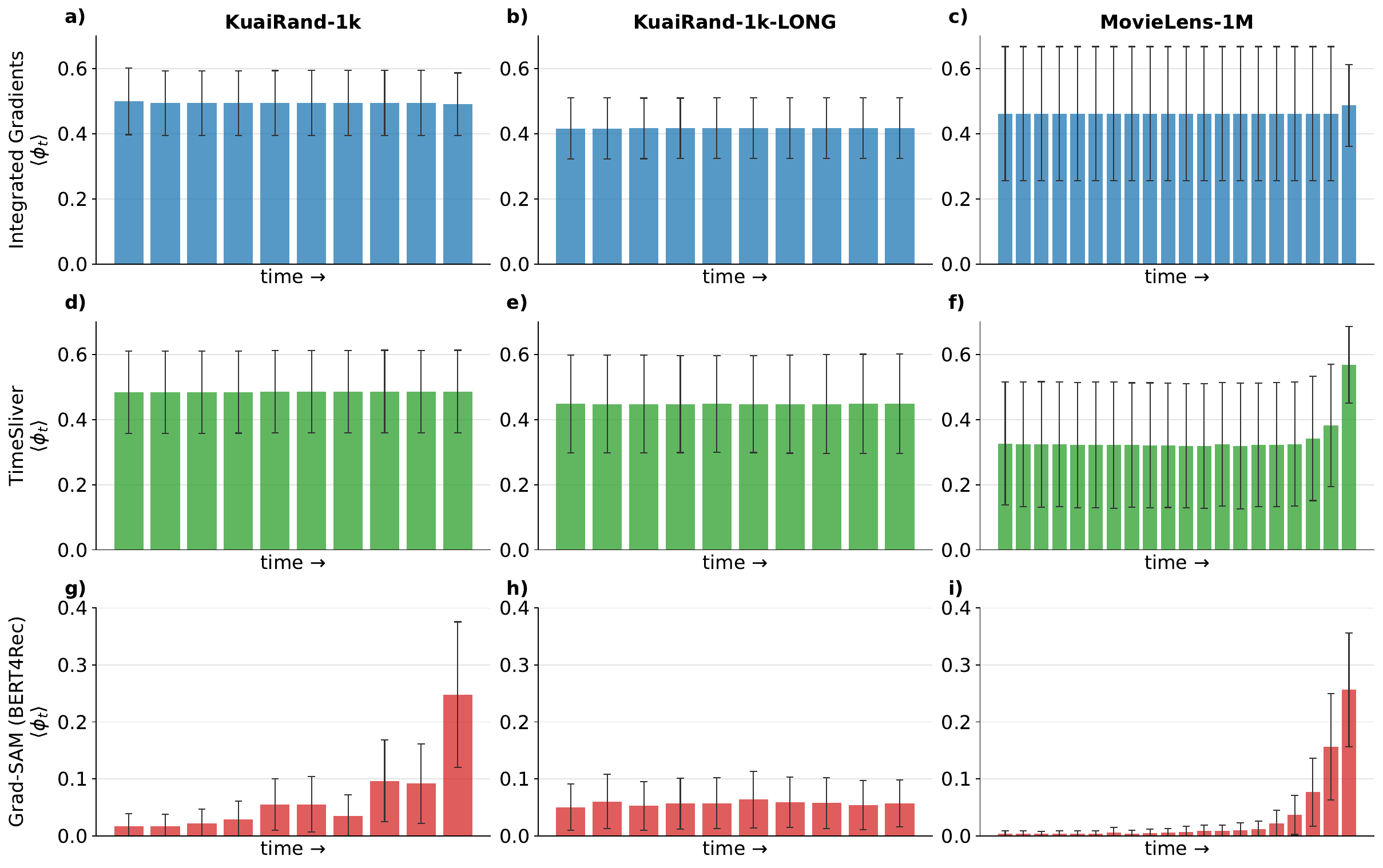}
  \vspace{-3mm}
  \caption{Temporal Importance Profiles of XAI Methods. Mean importance scores $<\phi_t>$ across the interaction sequence, for Integrated Gradients (a, b, c), TimeSliver (d,e,f), and Grad-SAM on BERT4Rec (g,h,i) for KuaiRand-1K, KuaiRand-1k-LONG, and MovieLens-1M datasets.}
\label{fig:recency}
\end{figure*}

\subsection{Attribution Mechanics: Recency and Popularity Bias}
  \label{sec:res-mechanics}

  \noindent\textbf{Are recent interactions more important?}  Figure~\ref{fig:recency} shows mean importance aggregated into ten-step bins for Integrated Gradients (\textit{CNN backbone}), TimeSliver, and Grad-SAM (BERT4Rec: \textit{Transformer backdone}). The temporal attribution pattern is  strongly dataset-dependent. On KuaiRand-1k, Integrated Gradients and TimeSliver produce flat profiles with no recency tilt, whereas on MovieLens-1M both methods concentrate
  importance on recent interactions — TimeSliver with a gradual monotonic ramp and Integrated Gradients with a pronounced spike on the final bin — consistent with next-item
  prediction being driven by the most recent items. Grad-SAM (BERT4Rec) exhibits notably different behavior across the two KuaiRand settings: on KuaiRand-1k it shows a sink-like
  concentration toward recent tokens \cite{xiao2024efficient}, while on KuaiRand-1k-LONG this pattern shifts substantially. This contrast is driven by the underlying softmax attention
  mechanism~\cite{mudarisov2025limitations}, which increasingly converges toward uniform token selection as sequence length grows, altering the attribution profile. While all three methods achieve $c_f < -0.8$, faithfulness alone does not fully characterize Grad-SAM: its reliance on the Transformer's self-attention mechanism can produce
  non-intuitive temporal attribution patterns that vary with sequence length, and practitioners should account for this before deploying it in production recommender systems. Full
   temporal profiles for all methods are provided in Appendix Figures~\ref{app_fig:kuai100}, \ref{app_fig:kuai1000}, and \ref{app_fig:ml}.

  \noindent\textbf{Are popular interactions rendered more important by XAI methods?}  Table~\ref{tab:freq_pcc} in Appendix \ref{app:detailed_results} reports the popularity bias score $c_p$ (Section~\ref{sec:method}) for all faithful methods ($c_f < -0.8$). Across all
  datasets, $c_p$ values remain within a narrow range of $[-0.146, +0.110]$, comparable to the random attribution baseline ($c_p = 0.007$), indicating
  that none of the evaluated methods exhibit a strong systematic bias toward popular items. This shows faithful attributions reflect true user preferences rather than item popularity.

% \begin{table}
% \caption{Frequency vs.\ Mean Importance: Pearson Correlation Coefficient (mean $\pm$ std over 5 folds of videos/items).}
% \label{tab:freq_pcc}
% \begin{tabular}{lccc}
% \toprule
%  & KuaiRand-1k(L = 100) & KuaiRand-1k (L = 1000) & MovieLens-1M \\
% Method &  &  &  \\
% \midrule
% Integrated Gradients & +0.011 ± 0.004 & -0.037 ± 0.009 & -0.061 ± 0.088 \\
% GradientSHAP & +0.008 ± 0.004 & -0.090 ± 0.008 & -0.033 ± 0.083 \\
% DeepLift & -0.009 ± 0.004 & -0.128 ± 0.008 & -0.026 ± 0.061 \\
% DeepLiftShap & -0.009 ± 0.004 & -0.128 ± 0.008 & -0.026 ± 0.061 \\
% KernelSHAP & +0.045 ± 0.005 & -0.146 ± 0.006 & -0.121 ± 0.083 \\
% Grad-SAM & -0.115 ± 0.003 & -0.001 ± 0.008 & -0.036 ± 0.121 \\
% TimeSliver & +0.110 ± 0.003 & -0.081 ± 0.005 & -0.123 ± 0.095 \\
% Grad-SAM (BERT4Rec) & — & — & -0.054 ± 0.039 \\
% Attn Tracing (BERT4Rec) & — & — & -0.513 ± 0.072 \\
% Grad-SAM (SASRec) & — & — & -0.085 ± 0.040 \\
% Attn Tracing (SASRec) & — & — & -0.467 ± 0.045 \\
% \bottomrule
% \end{tabular}
% \end{table}
\section{Limitations}
In this work, we provided a quantitative assessment of temporal attribution scores to evaluate whether datasets and models exhibit systematic recency or popularity biases. However, our focus on establishing a quantitative benchmarking framework leaves room for future user studies. Future work can validate these findings by conducting user studies to evaluate how technical attribution quality translates into practical explainability outcomes, domain-specific utility, and end-user perception in real-world systems.

\section{Conclusion}
In this work, we presented the first systematic benchmark of post-hoc XAI methods for sequential recommendation, evaluating ten attribution methods across CNN, Transformer,
  SASRec, and BERT4Rec backbones on KuaiRand-1k and MovieLens-1M. We introduced a dual-model faithfulness metric $c_f$ that decouples attribution from evaluation, avoiding the
  out-of-distribution pitfalls of naive masking. Our results show that gradient-based methods, particularly Integrated Gradients and GradientSHAP, are the most faithful and robust
   across all settings. Attention Tracing is highly architecture-dependent and unreliable as a faithfulness proxy, while gradient-weighted attention restores faithfulness on
  shorter sequences but degrades on longer horizons as softmax probabilities converge toward uniform importance scores. Our analysis further reveals that temporal attribution
  patterns in faithful methods reflect genuine task structure rather than recency or popularity bias, confirming that attribution quality is not confounded by sequence position or
   item frequency. Taken together, our findings provide actionable guidance for practitioners deploying explainable sequential recommenders: gradient-based post-hoc methods offer
  reliable and stable attributions, while attention-based methods should be adopted with caution and carefully evaluated against sequence length and architecture.

  Beyond post-hoc evaluation, our dual-model framework provides actionable insights for trajectory refinement and model efficiency in production systems. Specifically, identifying negative-impact or noisy interactions (such as accidental misclicks) enables targeted dataset denoising to improve sequence representations. Furthermore, these faithful attribution scores serve as supervisory signals for offline input sequence distillation, enabling online efficient inference.
% \vspace{-2mm}
% \begin{verbatim}
% \documentclass[sigconf, language=english, language=german,
%                language=french]{acmart}
% \end{verbatim}

% The title, subtitle, keywords and abstract will be typeset in the main
% language of the paper.  The commands \verb|\translatedXXX|, \verb|XXX|
% begin title, subtitle and keywords, can be used to set these elements
% in the other languages.  The environment \verb|translatedabstract| is
% used to set the translation of the abstract.  These commands and
% environment have a mandatory first argument: the language of the
% second argument.  See \verb|sample-sigconf-i13n.tex| file for examples
% of their usage.

\begin{acks}
We thank Giri Iyengar at Capital One for his support and sponsorship of this work.
\end{acks}

\bibliographystyle{ACM-Reference-Format}
\bibliography{recsys_xai}

\appendix

\section{Implementation Details}\label{app:impl}
\subsection{Dataset Details}

\begin{table*}[!htbp]
      \centering
      \caption{Dataset statistics.}
      \vspace{-3mm}
      \label{tab:datasets}
      \footnotesize
      \setlength{\tabcolsep}{5pt}
      \renewcommand{\arraystretch}{1}
      \begin{tabular}{lrrrrrrrr}
      \toprule
      \textbf{Dataset} & $L$ & \textbf{Users} & \textbf{Items} & \textbf{Sparsity} & $|\mathcal{D}_{train}|$ & $|\mathcal{D}_{valid}|$ & $|\mathcal{D}_{test}|$ \\
      \midrule
      KuaiRand-1k  & 100  & 100   & 456,664 & 98.68\% & 427,574 & 91,624 & 91,682 \\
      KuaiRand-1k-LONG  & 1000 & 100   & 2,664,050   & 99.76\%     & 378,818    & 81,178    & 81,235   \\
      MovieLens-1M & 200  & 6,040 & 3,706   & 95.53\% & 236,422 & 50,801 & 51,549 \\
      \bottomrule
      \end{tabular}
  \end{table*}

\subsection{Architectural Details}\label{app:arch_details}
All architectures use a learnable positional encoding added to the input embeddings prior to the main encoder.

\paragraph{CNN.}
The CNN model consists of a 2-layer MLP input projection mapping the input feature dimension to a hidden size of $128$, followed by three causal Conv1d
  layers with $128$ filters, kernel size $3$, ReLU activations, LayerNorm, and dropout ($p=0.1$). The output is aggregated via global average pooling and
  passed through a 2-layer MLP classifier.

\paragraph{Vanilla Transformer.}
The Transformer encoder consists of two weight-shared layers (ALBERT-style~\cite{lan2019albert}), each with multi-head self-attention ($4$ heads, hidden
dim $128$) and a position-wise feed-forward network (dimension $256$), with dropout ($p=0.1$). The sequence representation is obtained via last time point representation and fed into a 2-layer MLP classifier.

\paragraph{TimeSliver.}
Following~\cite{pandey2026timesliver}, the model projects input features to a dimension of $d=64$. A Conv1d layer with kernel size $1$ extracts
motif-level representations of dimension $16$, which are used to construct a temporal interaction heatmap via an outer product. The heatmap is reduced via
adaptive average pooling and passed to a linear classifier. Attribution scores are computed directly from the motif contributions, making the model
intrinsically interpretable.

\paragraph{SASRec.}
  We use the standard SASRec architecture~\cite{kang2018sasrec} with $2$ self-attention blocks, embedding dimension $128$, $4$ attention heads, and dropout
  rate $0.2$. SASRec is trained on MovieLens-1M only.

  \paragraph{BERT4Rec.}
  We use the standard BERT4Rec architecture~\cite{sun2019bert4rec} with $2$ Transformer layers, embedding dimension $128$, $4$ attention heads, dropout rate
   $0.2$, and a masked item prediction rate of $15\%$ during training. BERT4Rec is trained on MovieLens-1M only.

\begin{table*}[!htbp]
    \centering
    \caption{Number of trainable parameters per architecture and dataset.}
    \label{app_tab:num_parameters}
    \begin{tabular}{llrrr}
    \toprule
    \textbf{Architecture} & \textbf{Variant}
        & \textbf{KuaiRand-1k}
        & \textbf{KuaiRand-1k-LONG}
        & \textbf{MovieLens-1M} \\
    \midrule
    CNN          & Vanilla   & 480K  & 467K & 257K    \\
    \multirow{3}{*}{Transformer}
        & Vanilla   & 464K  & 451K & 241K    \\
        & SASRec    & 563K   & 678K & 728K \\
        & BERT4Rec  & 563K   & 678K & 748K \\
    TimeSliver   & Vanilla   & 71K   & 70K & 26K     \\
    \bottomrule
    \end{tabular}
  \end{table*}
  
\subsection{XAI Method Implementation}\label{app:xai_impl}

All gradient-based and perturbation-based methods use a zero tensor as the reference baseline, ensuring that ``absence'' of an interaction is represented
  consistently across attribution, masking, and evaluation. Attribution scores over the embedding dimension are aggregated to a per-timestep scalar by
  taking the $L_2$ norm across the feature axis.

  \paragraph{Integrated Gradients (IG)}
  Attributions are computed using $20$ integration steps along the straight-line path from the zero baseline to the input embedding.

  \paragraph{GradientSHAP}
  Shapley values are approximated using $20$ gradient samples with a zero baseline distribution of $4$ reference samples and no additional input noise
  ($\text{stdevs}=0$). Random seeds are fixed immediately before the attribution call to ensure reproducibility.

  \paragraph{DeepLIFT}
  Contribution scores are propagated by comparing neuron activations to a zero reference baseline.

  \paragraph{DeepLIFT-SHAP}
  Extends DeepLIFT by averaging over a baseline distribution of $3$ near-zero reference samples ($\mathcal{N}(0, 10^{-4})$). Random seeds are fixed before
  the attribution call.

  \paragraph{Grad-CAM}
  Forward and backward hooks are registered on the final convolutional layer. The per-channel gradient means are used to weight the activation maps, which
  are summed across channels and passed through ReLU. Scores are min-max normalised per sample.

  \paragraph{KernelSHAP and LIME}
  Both methods operate on a binary mask representation $\mathbf{z} \in \{0,1\}^L$, where $z_t = 0$ replaces the embedding at timestep $t$ with the zero
  baseline. Feature masks assign one group per timestep, so each attribution directly corresponds to a single interaction. A batch size of $32$ is used for
  tractability.

  \paragraph{Attention Tracing}
  Attention weights are rolled out across all layers and heads using the last sequence position as the query, following~\cite{attentiontracing}.

  \paragraph{Grad-SAM}
  Attention maps from all heads and layers are reweighted by the gradient of the target output with respect to the attention scores,
  following~\cite{gradsam}.

  \paragraph{TimeSliver.}
  Attribution scores are computed natively from the model's motif-level gradient decomposition, mapping directly to per-timestep importance without any
  post-hoc aggregation. Scores are normalised per sample by the maximum absolute value.

% \subsection{Code Snippets for Reproducibility}\label{app:code_repo}
% \begin{figure}[H]
%   \centering
%  \includegraphics[width=0.9\linewidth]{figs/captum.png}
%   \caption{Code snippet for Integrated Gradient, GradientSHAP, DeepLift and DeepLiftShap XAI methods.}
% \end{figure}

% \begin{figure}[H]
%   \centering
%  \includegraphics[width=0.9\linewidth]{figs/lime_ks.png}
%   \caption{Code snippet for LIME and KernelSHAP XAI method.}
% \end{figure}

% \begin{figure}[H]
%   \centering
%  \includegraphics[width=0.9\linewidth]{figs/gradcam.png}
%   \caption{Code snippet for GradCAM XAI method.}
% \end{figure}

% \begin{figure}[H]
%   \centering
%  \includegraphics[width=0.9\linewidth]{figs/attention_tracing.png}
%   \caption{Code snippet for Attention tracing XAI method.}
% \end{figure}

% \begin{figure}[H]
%   \centering
%  \includegraphics[width=0.9\linewidth]{figs/gradsam.png}
%   \caption{Code snippet for GradSAM XAI method.}
% \end{figure}

\section{Detailed Results}\label{app:detailed_results}
\begin{table}[!htbp]
  \centering
  \caption{Impact of maximum sequence length $L$ on the faithfulness correlation metric $c_f$ (reported as mean $\pm$ std) for Grad-SAM XAI method.}
  \label{app:tab:gradsam_length}
  \renewcommand{\arraystretch}{1.1} % Comfortable row padding
  \resizebox{\linewidth}{!}{%
  \begin{tabular}{cc}
  \toprule
  \textbf{Sequence Length ($L$)} & \textbf{Faithfulness Correlation ($c_f$)} \\
  \midrule
  50   & $-0.74559 \pm 0.0286$ \\
  100  & $-0.82500 \pm 0.0550$ \\
  200  & $-0.80780 \pm 0.0817$ \\
  300  & $-0.94590 \pm 0.0261$ \\
  500  & $-0.36800 \pm 0.2259$ \\
  750  & $-0.66600 \pm 0.0955$ \\
  1000 & $-0.30770 \pm 0.1546$ \\
  \bottomrule
  \end{tabular}}
\end{table}

  \begin{table*}[!htbp]
    \centering
    \caption{Predictive performance of backbone architectures. AUC-ROC for KuaiRand-1k (binary classification) and NDCG@10 for MovieLens-1M (next-item
  prediction). Results are mean $\pm$ std over three random seeds.}
    \label{tab:predictive}
    \setlength{\tabcolsep}{8pt}
    \begin{tabular}{llccc}
    \toprule
    \textbf{Architecture} & \textbf{Variant}
        & \shortstack[c]{\textbf{KuaiRand-1k} \\ \textbf{($L=100$, AUC-ROC)}}
        & \shortstack[c]{\textbf{KuaiRand-1k} \\ \textbf{($L=1000$, AUC-ROC)}}
        & \shortstack[c]{\textbf{MovieLens-1M} \\ \textbf{(NDCG@10)}} \\
    \midrule
    CNN          & ---       & $0.758 \pm 0.015$ & $0.7198 \pm 0.0212$ & $0.4109 \pm 0.0023$ \\
    \midrule
    \multirow{3}{*}{Transformer}
        & Vanilla   & $0.761 \pm 0.013$ & $0.7322 \pm 0.0200$ & $0.4148 \pm 0.0040$ \\
        & SASRec    & $0.767 \pm 0.034$               & $0.759 \pm 0.001$ & $0.4209 \pm 0.0048$ \\
        & BERT4Rec  & $0.778 \pm 0.022$               & $0.749 \pm 0.001$ & $0.4310 \pm 0.0030$ \\
    \midrule
    TimeSliver   & ---       & $0.758 \pm 0.015$ & $0.7216 \pm 0.0218$ & $0.3665 \pm 0.0008$ \\
    \bottomrule
    \end{tabular}
  \end{table*}

% \begin{table}[!htbp]
%   \centering
%   \caption{Masking faithfulness evaluation: Pearson correlation (mean $\pm$ std over
%   5 segments) between importance rank and drop in $P(\text{true item})$ under
%   progressive masking, on MovieLens-1M (Classification). Values are averaged across seeds 40, 41, and 42.
%   Higher magnitude negative values indicate stronger faithfulness.}
%   \label{tab:faithfulness}
%   \setlength{\tabcolsep}{12pt}
%   \begin{tabular}{llc}
%   \toprule
%   \textbf{Architecture} & \textbf{Method} & \shortstack[c]{\textbf{MovieLens-1M} \\ \textbf{Classification}} \\
%   \midrule
%   Baseline
%       & Random
%       & $-0.101 \pm 0.181$ \\
%   \midrule
%   \multirow{7}{*}{CNN}
%       & Integrated Gradients
%       & $+0.949 \pm 0.005$ \\
%       & GradientSHAP
%       & $+0.939 \pm 0.006$ \\
%       & DeepLift
%       & $+0.993 \pm 0.001$ \\
%       & DeepLiftSHAP
%       & $+0.993 \pm 0.001$ \\
%       & KernelSHAP
%       & $+0.945 \pm 0.006$ \\
%       & LIME
%       & $+0.838 \pm 0.011$ \\
%       & Grad-CAM
%       & $-0.796 \pm 0.016$ \\
%   \midrule
%   \multirow{2}{*}{Transformer}
%       & Attention Tracing
%       & $-0.871 \pm 0.033$ \\
%       & Grad-SAM
%       & \textbf{$-0.899 \pm 0.003$} \\
%   \midrule
%   TimeSliver
%       & TimeSliver
%       & \underline{$-0.888 \pm 0.009$} \\
%   \bottomrule
%   \end{tabular}
% \end{table}

%%%% detailed temporal importance scores
\clearpage
\begin{figure*}[t]
  \centering
 \includegraphics[width=0.9\linewidth]{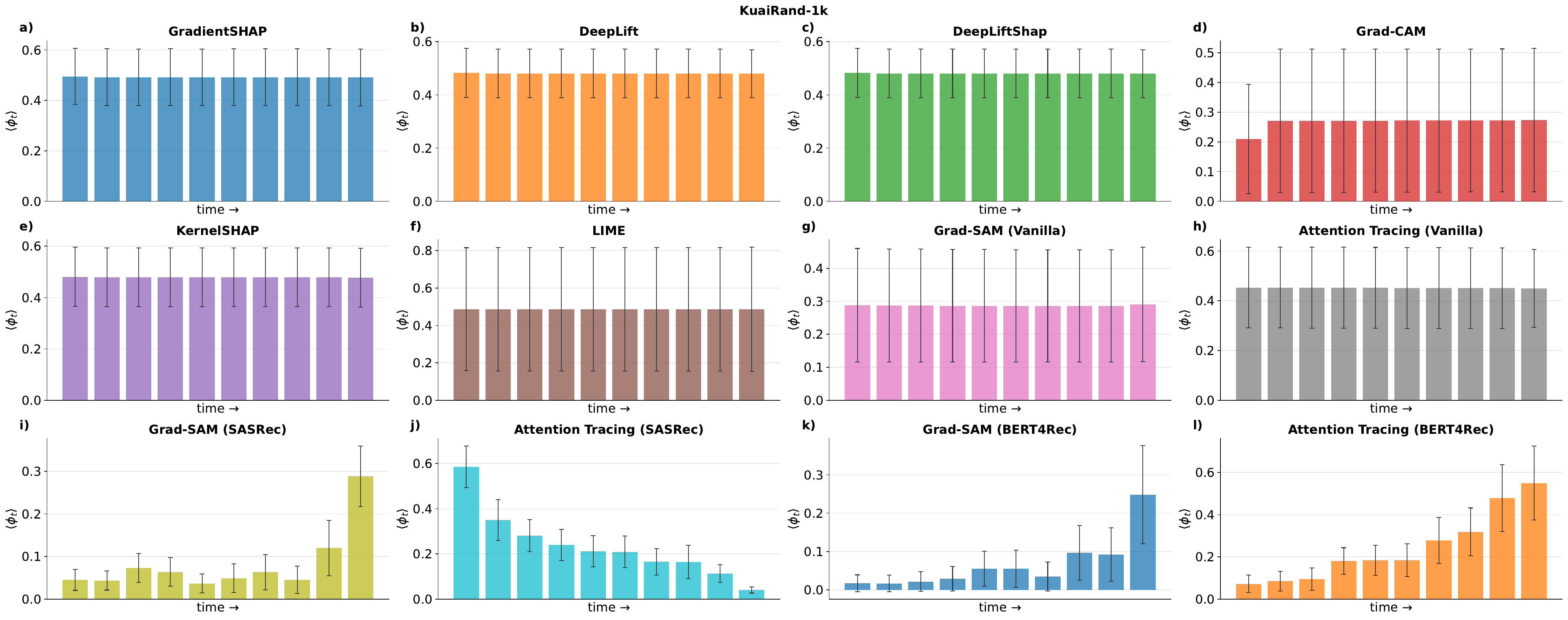}
  \caption{Temporal Importance Profiles of XAI Methods for KuaiRand-1k dataset with L=100. Mean importance scores aggregated into 10-step time bins across the interaction sequence. Error bars denote ±1 standard deviation. The x-axis progresses from the oldest interaction (left) to the most recent (right).}
\label{app_fig:kuai100}
\end{figure*}

\begin{figure*}[t]
  \centering
 \includegraphics[width=0.9\linewidth]{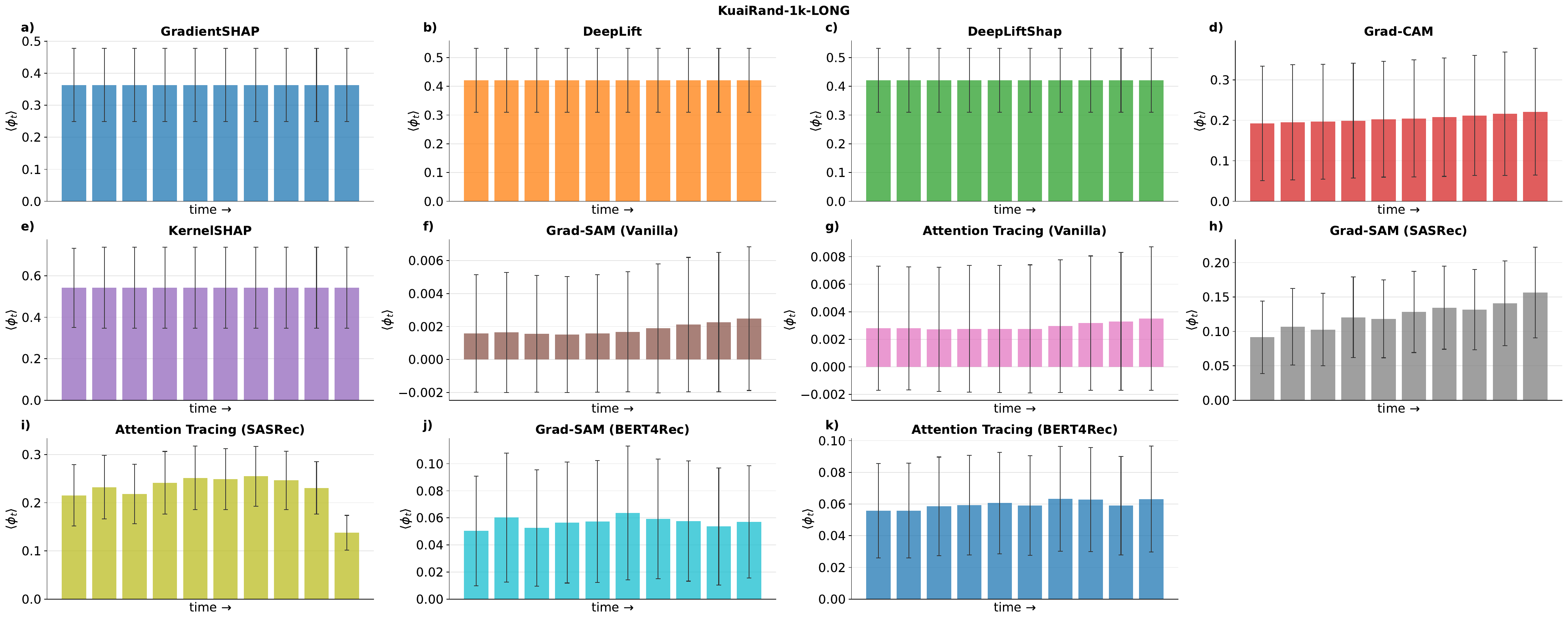}
  \caption{Temporal Importance Profiles of XAI Methods for KuaiRand-1k-LONG dataset with L=1000. Mean importance scores aggregated into 10-step time bins across the interaction sequence. Error bars denote ±1 standard deviation. The x-axis progresses from the oldest interaction (left) to the most recent (right).}
\label{app_fig:kuai1000}
\end{figure*}

\clearpage
\begin{figure*}[t]
  \centering
 \includegraphics[width=0.9\linewidth]{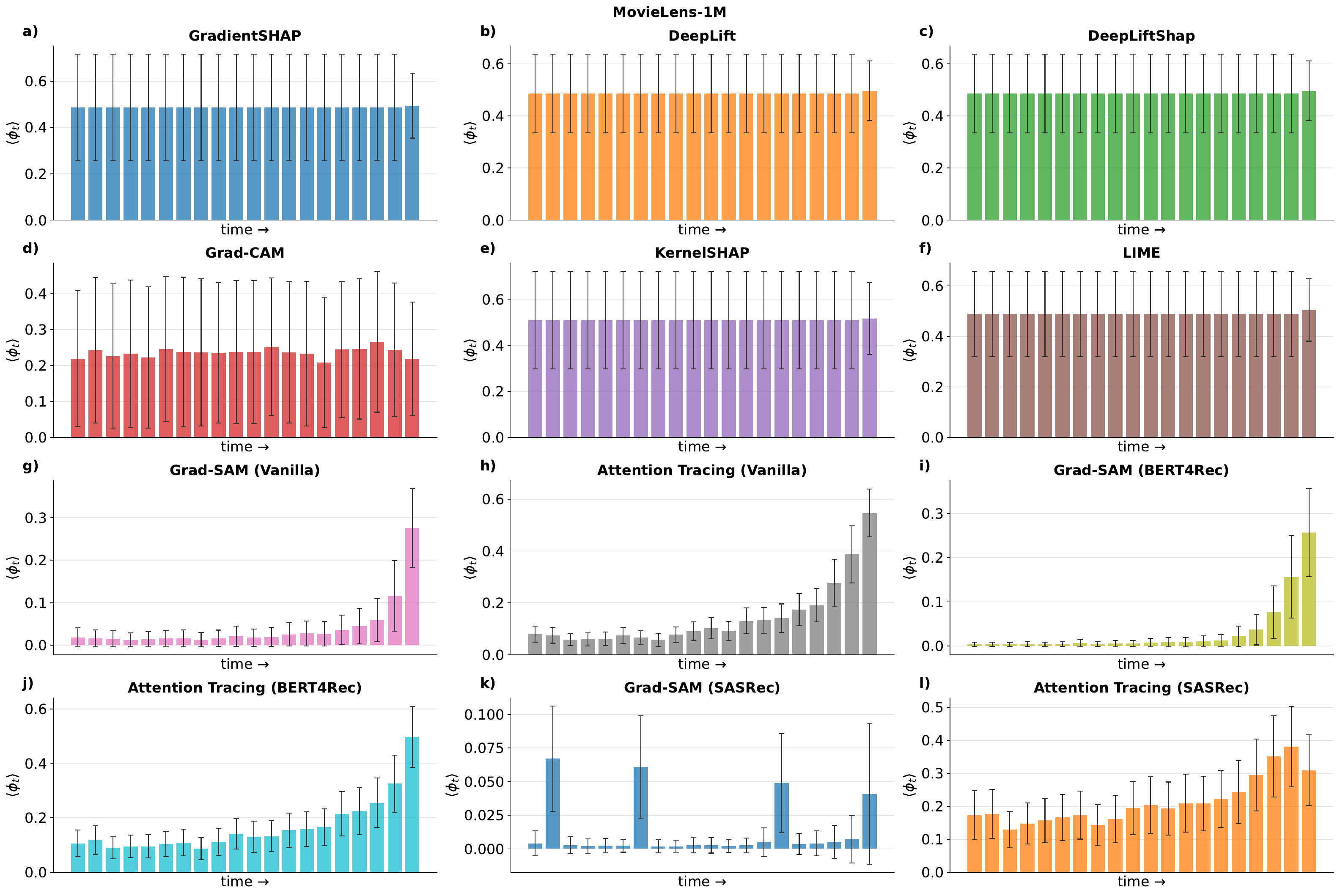}
  \caption{Temporal Importance Profiles of XAI Methods for MovieLens-1M dataset. Mean importance scores aggregated into 10-step time bins across the interaction sequence. Error bars denote ±1 standard deviation. The x-axis progresses from the oldest interaction (left) to the most recent (right).}
\label{app_fig:ml}
\end{figure*}

\begin{table*}[!htbp]
      \centering
      \caption{Popularity bias score $c_p$ (mean $\pm$ std over 5 folds). The lowest value is highlighted in \textcolor{forestgreen}{green} and the
  highest in \textcolor{maroon}{maroon}. \textbf{Higher values are not favorable}.}
      % \vspace{-4mm}
      \label{tab:freq_pcc}
      \footnotesize
      \setlength{\tabcolsep}{4pt}
      \renewcommand{\arraystretch}{0.9}
      \begin{tabular}{llccc}
      \toprule
      \textbf{Method} & \textbf{Variant} & \textbf{KuaiRand-1k} & \textbf{KuaiRand-1k-LONG} & \textbf{MovieLens-1M} \\
      \midrule
      Integrated Gradients & ---        & $+0.011 \pm 0.004$ & $-0.037 \pm 0.009$ & $-0.061 \pm 0.088$ \\
      GradientSHAP         & ---        & $+0.008 \pm 0.004$ & $-0.090 \pm 0.008$ & $-0.033 \pm 0.083$ \\
      DeepLift             & ---        & $-0.009 \pm 0.004$ & $-0.128 \pm 0.008$ & \textcolor{maroon}{$-0.026 \pm 0.061$} \\
      DeepLiftSHAP         & ---        & $-0.009 \pm 0.004$ & $-0.128 \pm 0.008$ & \textcolor{maroon}{$-0.026 \pm 0.061$} \\
      KernelSHAP           & ---        & $+0.045 \pm 0.005$ & \textcolor{forestgreen}{$-0.146 \pm 0.006$} & $-0.121 \pm 0.083$ \\
      TimeSliver           & ---        & \textcolor{maroon}{$+0.110 \pm 0.003$} & $-0.081 \pm 0.005$ & \textcolor{forestgreen}{$-0.123 \pm 0.095$} \\
      \midrule
      \multirow{3}{*}{Grad-SAM}
          & Vanilla    & \textcolor{forestgreen}{$-0.115 \pm 0.003$} & \textcolor{maroon}{$-0.001 \pm 0.008$} & $-0.036 \pm 0.121$ \\
          & SASRec     & $-$ & $-$ & $-0.085 \pm 0.040$ \\
          & BERT4Rec   & $-$ & $-$ & $-0.054 \pm 0.039$ \\
      \midrule
      \end{tabular}
  \label{tab:popularity}
  \end{table*}

\end{document}